\documentclass[10pt, conference, compsocconf]{IEEEtran}
\IEEEoverridecommandlockouts

\usepackage{times}
\usepackage{epsfig}
\usepackage{graphicx}
\usepackage{amsmath}
\usepackage{amssymb}

\usepackage[utf8]{inputenc}
\usepackage{mathtools}
\usepackage{float}
\usepackage{amsthm}
\usepackage{thmtools}
\usepackage{commath}
\usepackage{blindtext}
\let\labelindent\relax
\usepackage{enumitem}
\usepackage{xcolor}
\usepackage{subfig}
\usepackage{changepage}
\usepackage{etoolbox}

\usepackage{wrapfig}

\newtheorem{lemma}{Lemma}

\usepackage[pagebackref=true,breaklinks=true,letterpaper=true,colorlinks,bookmarks=false]{hyperref}

\begin{document}
\title{Towards Safe, Real-Time Systems: Stereo vs Images and LiDAR for 3D Object Detection}

\author{\IEEEauthorblockN{Matthew Levine$^{1}$} \thanks{Matthew Levine is with the National Robotics Engineering Center, Carnegie Mellon University, Pittsburgh, PA 15201  {\tt\small mjlevine@andrew.cmu.edu}}}

\maketitle

\begin{abstract}
 As object detectors rapidly improve, attention has expanded past image-only networks to include a range of 3D and multimodal frameworks, especially ones that incorporate LiDAR. However, due to cost, logistics, and even some safety considerations, stereo can be an appealing alternative. Towards understanding the efficacy of stereo as a replacement for monocular input or LiDAR in object detectors, we show that multimodal learning with traditional disparity algorithms can improve image-based results without increasing the number of parameters, and that learning over stereo error can impart similar 3D localization power to LiDAR in certain contexts. Furthermore, doing so also has calibration benefits with respect to image-only methods. We benchmark on the public dataset KITTI, and in doing so, reveal a few small but common algorithmic mistakes currently used in computing metrics on that set, and offer efficient, provably correct alternatives.
\end{abstract}

\IEEEpeerreviewmaketitle

\section{Introduction}

Object detection in popular domains like automotive has a deep bench of solutions. Single-shot detectors provide real-time, embedded opportunities \cite{womg2018tiny}, while two-stage detectors support highly accurate models \cite{ren2016faster}. Fundamentally differentiating these approaches - and separating them from preceding traditional approaches - is the proposal scheme, which dictates the regions-of-interest in the input space to be classified and regressed over. The flexibility of these proposal methodologies combined with the fundamental power of convolutional frameworks in image domains provides for robust potential when optimizing traditional, 2D localization.

Modern deep learning goes further. Although developed to work with images, CNNs have attempted to move from image-based input to 3D inputs like LiDAR (e.g. \cite{he2020structure}, \cite{ali2018yolo3d}).  A corollary advancement predicts object locations in 3D, a vital objective for safety systems. These methodologies are not mutually exclusive in the literature, with some attempts to fuse LiDAR with camera data, e.g., by leveraging the 3D inputs for anchor generation and the the image features for box regression and classification \cite{furst2020lrpd}. 

In some cases, it may be desirable to utilize stereo derived from two calibrated cameras instead of LiDAR. Often, this motivation is as simple as cost or the logistical hurdle of incorporating and synchronizing LiDAR with cameras; stereo is many times cheaper, and naturally is produced in the same space as the image pairs. While there are clear safety benefits to adding a LiDAR to an autonomy system \cite{rasshofer2005automotive}, these can be undermined if a model learns jointly over a LiDAR and image. In that case, sensor redundancy can be lost and the model may fail if either inputs undergo degradation. If stereo can be substituted for other range sensors, it frees them up to be used for independent object detection towards a redundancy-driven safety case. 

We characterize the trade-off of utilizing stereo data by empirically bounding the performance of stereo methods, underneath by one-to-one image-only methods, and above by state-of-the-art LiDAR methods:

\begin{itemize}
  \item Towards constrained systems, we examine singleshot frameworks that are efficient - relying only on block-matching for stereo inference - and improve on equivalent image-only networks when trained with the exact same procedure and given the same number of parameters; AP increases from 0.605 to 0.907.
  \item Suggesting that error in stereo can be explicitly learned over and that stereo can be a plausible substitute for LiDAR in some settings, we show that the $l_2$ localization difference between stereo and LiDAR models is generally between $0.04$m and $0.1$ms on KITTI
  \item We show that deep multimodal learning with independently computed stereo imbues some calibration invariance versus monocular models, and examine this stability under sensor degradation 
\end{itemize}

Towards evaluating these models for safety, we also make small note of some common metric deficiencies. The \textit{true positive rate} is unambiguous in classification schemes, but in object detection requires pairing detections and labels accurately. When multiple labels and detections for a single input are not disjoint, this pairing is nontrivial. Existing greedy algorithms, like the ones used by default in VOC toolkit and KITTI, can make infrequent pairing errors in the case of overlapping detections. Furthermore, their metric computation can be unstable or lack coherency when filtering for subset criteria - like they do with the difficulty classes - which may be necessary when building a safety case. We conclude with a brief discussion of these deficiencies and recommended remedy.

\section{Related Work}

Although less common than monocular or LiDAR networks, some authors already use stereo for object detection. One popular method involves inputting the left and right images jointly, to implicitly learn stereo information \cite{li2019stereo} \cite{chen2020dsgn} \cite{sun2020disp}. While performant, this methodology discards known information about the extrinsics of the cameras, which may introduce problems if these settings differ at evaluation (e.g., if the cameras' imagers or their orientation and position can change, as is common in cheaper plug-and-play or self-mounted systems).  Because any depth map generated is done implicitly, it is more difficult to build a secondary geometry-based detector to fuse with the image detector for security-driven redundancy, as is common in robotics applications (e.g., \cite{stentz2002system}). Furthermore, it doubles the number of parameters necessary for input to the network, which may pose speed challenges in real-time systems or memory challenges when deploying on memory-sensitive hardware like FPGAs. 

Another method involves using disparity images, but only for anchor generation \cite{konigshof2019realtime}. This model is not fully learned, and may be undermined at the clustering stage; while it may be effective for distinct cases, heavily occluded objects may be difficult to segregate. The authors of \cite{chen20173d} compare methods of involving stereo, though they use a complicated stereo algorithm that requires segmentation and scene-flow analysis; moreover, however, their approach is centered on incorporating stereo into a cohesive 3D region proposal scheme. It requires HHA features \cite{gupta2014learning}, which also doubles the number of network parameters. Requiring manually engineered features runs counter to the current trend of allowing the model to learn end-to-end without hand-tuning.

The authors of \cite{wang2019pseudo} directly apply stereo information into LiDAR systems without attempting to fuse image information, and characterize the difference exclusively in terms of AP, which may hide in aggregate some of the particular failure surfaces relevant to a safety case involving stereo data.

\begin{table}
\begin{center}
\begin{tabular}{|l|c|c|}
\hline
Method & A.P. & Brier \\
\hline\hline
SSD-SA (LiDAR) & \textbf{0.986} & 0.132 \\
S-SSD (Stereo) & 0.852 & 0.222 \\
SSD550 (RGB) & 0.605 & 0.344 \\
SSD550 (RGB-H) & \textbf{0.907} & \textbf{0.110}\\
RPN (RGB) & 0.871 & 0.117 \\
RPN (RGB-D) & \textbf{0.940} & \textbf{0.061} \\  
\hline
SSD550 (RGB) $\sigma=5$ & 0.019 & 0.949 \\
SSD550 (RGB-H) $\sigma=5$ & 0.551 & 0.970\\
RPN (RGB) $\sigma=5$ & 0.440 & 0.568\\
RPN (RGB-D) $\sigma=5$ & 0.482 & 0.593\\
\hline
\end{tabular}
\end{center}
\caption{AP and Brier Score for models (since Brier is an $l_2$ distance to the ideal calibration curve, note that a lower Brier score is desirable.). IoU of 0.7 required for accurate matches. Metrics given for "Medium" difficulty class on KITTI, in keeping with the leaderboard's recommendation. The metrics from models existing in the literature (SSD-SA LiDAR, SSD, RPN) are slightly different than they appear reported on the KITTI leaderboard since they were retrained and evaluated on an isolated partition of the training set.}
\label{table:table1Performance}
\end{table}

\section{Methods}
Our analysis considers techniques that (a) maintain speed parity with existing single-shot models while improving performance and calibration and (b) understand the maximal benefit available in stereo information.

\subsection{Single-Shot, 1-1 Image Improvement} 
\begin{figure}[!htbp]
\begin{center}
   \includegraphics[width=1.0\linewidth]{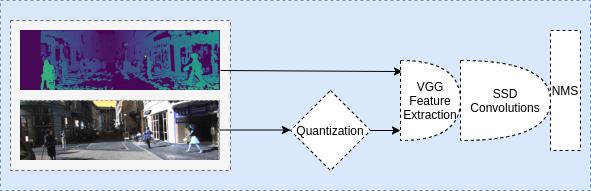}
\end{center}
   \caption{Network architecture for RGB + Height (H) features. The inputs are projected disparity information from stereo and quantized pixel data; an SSD architecture follows}
\label{fig:networkDesign1}
\end{figure}

Standard image representation encodes three channels in 8-bits each. To preserve the number of parameters exactly, we quantize the image from 8 bits-per-channel to 5, leaving 9 bits of the originally allocated space for geometric information. We then concatenate an additional channel with a height image taken from the reprojection of disparity and normalized into the pixel domain. Unlike a bird's eye input used in some region-proposal schemes, the disparity and height image are in the same frame of reference as the left image. Since the only operational cost is the block matching expedition and a reprojection, the overhead is trivial compared to the network speed. This network is shown in Figure \ref{fig:networkDesign1}.

The image channels can be compressed with an out-of-the-box algorithm that assigns weights to the components as in a color conversion or debayering kernel. However, in doing so \footnote{performed here using algorithms from the OpenCV API}, an SSD model's average precision on KITTI improves only modestly this way, from AP of $0.605$ to $0.621$. On the other hand, we can jointly learn the compression weighting with the model to find the optimal quantized representation by introducing an additional convolution up-front. Although this could be reasonably characterized as increasing the number of parameters of the network (and it certainly is, during training), we think of this as learning a pre-processing step equivalent to running a compression kernel; in any case, the kernel size is very small compared to the backbone. In this scheme, as we see in Table \ref{table:table1Performance}, AP improves dramatically, from $0.605$ to $0.907$, and results from this method are presented in subsequent sections.

\subsection{Upper-bound Improvement on Image Detectors}

\begin{figure}[H]
\begin{center}
   \includegraphics[width=1.0\linewidth]{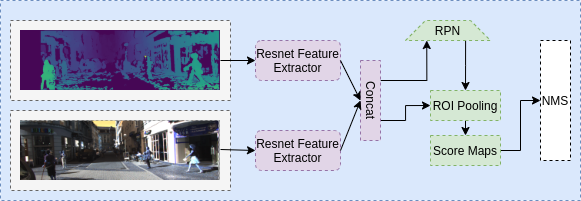}
\end{center}
   \caption{Network architecture for Stereo RPN.}
\label{fig:networkDesign2}
\end{figure}

On the other hand, Figure \ref{fig:networkDesign2} shows a more expensive version of multimodal learning that optimizes for performance over speed using a late-fusion Resnet-based RPN. This network passes both height and depth instead of using raw disparity images, effectively mimicking a 3D pointcloud in the plane of the image. The separate channel of convolutions removes any continuity between the spaces that a single-channel approach might imply. Moreover, the two-channel approach allows mirroring pretrained weights while allowing for separate feature kernels, potentially avoiding loss valleys. As opposed to previous work like \cite{chen20173d}, this method focuses on feature extraction from the stereo input, not region proposal and is presented here not as a novel methodology, but to understand to comparative performance of the different stereo options. This method is referred to as RPN with RGBD information, to distinguish it from the image-only RPN with RGB features.

\subsection{Model Evaluation}

Some authors caution that models in safety cases need to be evaluated beyond their nominal cases \cite{pezzementi2018putting}. Towards understanding the different architecture benefits, we measure performance on some of the models under image degradation; in particular, we apply Gaussian filters to the input image at evaluation (but not training) to model defocus. To quantify calibration, in addition to performance, we provide a Brier score, though because this value is highly sensitive to the underlying support, there are additional notes in how it should be computed in the final section.

We provide results on the KITTI object detection benchmark, focusing on ROC curves, AP, and Brier score as a measure of calibration. We default to the "Medium" difficulty KITTI subset except where specified otherwise.

In comparing the calibration of the networks, we also demonstrate an auxiliary benefit to RGB-H input over images; under certain degredations, performance in image-only network appears to collapse much faster than in stereo-augmented networks.

\subsection{Stereo vs LiDAR}

To expand the case that stereo can be a situational replacement for LiDAR and quantify the error implicit in learning over block matching, we directly compare state-of-the-art LiDAR methods in SSD-SA\cite{he2020structure} by adapting them to stereo point clouds generated from block matching. 

The SSD-SA model is a single-stage 3D object detector leveraging several key insights, many of which do not require modification when using stereo input. Part-sensitive warping for improved calibration, multi-resolution features, and point-wise supervisors can be directly extrapolated to the new sensor. As opposed to schemes that voxelize the point cloud, SSD-SA maps points directly into indices in the input. For stereo data, a simple pre-processing that removes invalid points and truncates by distance is substituted. Where SSD-SA discards LiDAR coordinates not represented in the image view during training, this is unnecessary when using stereo since those points cannot exist. We accept the cut-and-paste augmentations that are used in the original LiDAR settings for this work.

\section{Results}

\begin{figure}[!htbp]
\centering
\includegraphics[width=0.8\linewidth]{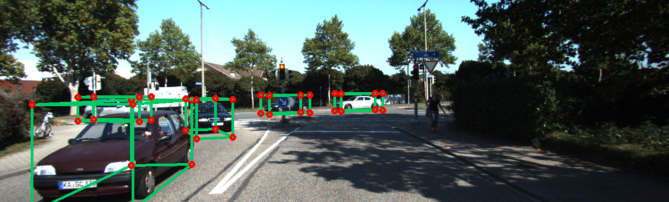}
\caption{From stereo-only 3D SSD model. The model overcomes stereo noise to detect occluded objects, with degraded localization for the closest and farthest objects}
\label{fig:rocComp1}
\end{figure}

\subsection{Stereo v. LiDAR for 3D Localization}

The ROC curve in Figure \ref{fig:rocComp1} shows that stereo-only approaches perform competitively with LiDAR both overall and on the reduced "medium" difficulty vehicles; the tail of accuracy for stereo is shorter, especially when considering difficult objects which tend to be heavily occluded or far away, so may be susceptible to stereo noise. In particular, from Table \ref{table:table1Performance}, we see that AP degrades from $0.986$ to $0.852$ when switching from LiDAR to stereo point clouds.   

\begin{figure}[!htbp]
\centering
\includegraphics[width=0.8\linewidth]{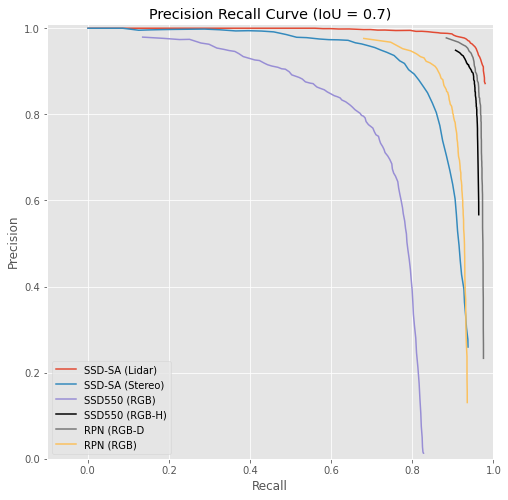}
\caption{ROC curve showing comparative performance between LiDAR, Stereo, and image models on Medium difficulty KITTI vehicles with IoU of 0.7. Notable is the very large performance gap between image-only models and those that use stereo}
\label{fig:rocComp2}
\end{figure}

\begin{figure}[!htbp]
\centering
\includegraphics[width=0.8\linewidth]{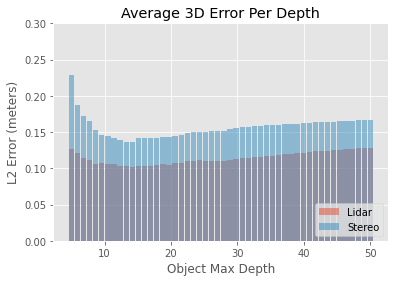}
\includegraphics[width=0.8\linewidth]{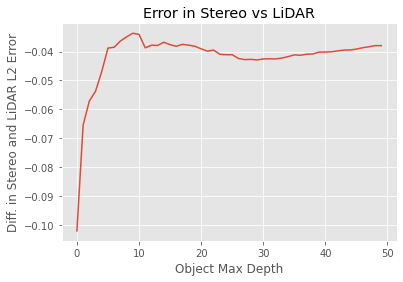}
\caption{(Top) Comparison in l2 3D error between Stereo and LiDAR models at various distances. (Bottom) Difference in $l_2$ 3D error between Stereo and LiDAR models and various distances}
\label{fig:l2errorDiff}
\end{figure}

However, we see in figure \ref{fig:l2errorDiff} that although the error for the stereo model is highest for close objects less than $8m$ away (as implied by the wide baseline of the KITTI cameras) and increases gradually in error after $15m$, the overall difference in the models' depth prediction differs by between $0.04$ and $0.1$ meters. For many applications, including slower moving vehicles in constrained environments, this may be a tolerable error considering the difference in price between the sensors. This error is also significantly lower than may be expected by block matching-based disparity algorithms; at 10m, we would expect about $0.28m$ error\footnote{from the standard derivation that $\nabla z = \dfrac{z^2}{b\cdot f}\nabla D$}, and this error should increase quadratically with distance. Thus, we appear to have been able to learn over the intrinsic error in the disparity computation, to a limited extent.

\begin{figure}[!htbp]
\centering
\includegraphics[width=0.8\linewidth]{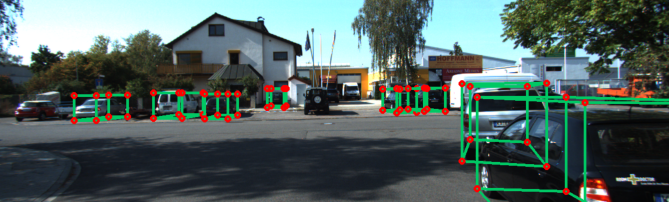}
\includegraphics[width=0.8\linewidth]{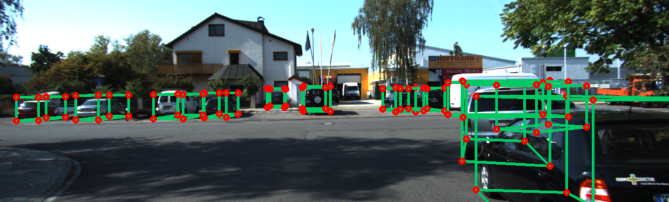}
\caption{(Top) stereo SSD misses some objects far away, especially when they are on the edges of the image or in low-texture shadows. (Bottom) LiDAR SSD can sometimes detect these corner cases. In all evaluations, we collapse Van and Car categories to ignore cross-class error.}
\label{fig:rocComp3}
\end{figure}

We note that machine-learning disparity algorithms, while widely better for many tasks, did not improve performance when substituted for BM, so are not reported here. Outputs from \cite{yin2019hierarchical}, e.g., yielded no significant change in AP; although it produces smoother disparity, it could  be that "double-learning" over the images, as happens when learned once to produce disparity and then again for object detection, provides no statistical benefit. Learning depth maps and object detection jointly may still provide improvements as an auxiliary-task scheme.

\subsection{Stereo to Augment Image Detectors}

Incorporating stereo in image-driven single-shot detection provides moderate improvements in AP. On the one hand, even keeping the number of parameters the same, we see in Table \ref{table:table1Performance} that adding stereo features improves AP from $0.605$ to $0.907$ when requiring 0.7 IoU for overlaping boxes. This even outperforms a Resnet-backboned image-only RPN, which has $AP=0.871$ and significantly more parameters. A full RPN with depth features, on the other hand, approaches SSD-SA's $0.986$ benchmark, achieving $0.951$ AP. Furthermore, stereo-based approaches improved calibration, bringing the Brier score for SSD from $0.344$ to $0.110$. The effect was even more dramatic with Resnet-RPN, with Brier improving from $0.117$ to $0.061$.

\begin{figure}[!htbp]
\centering
\includegraphics[width=0.8\linewidth]{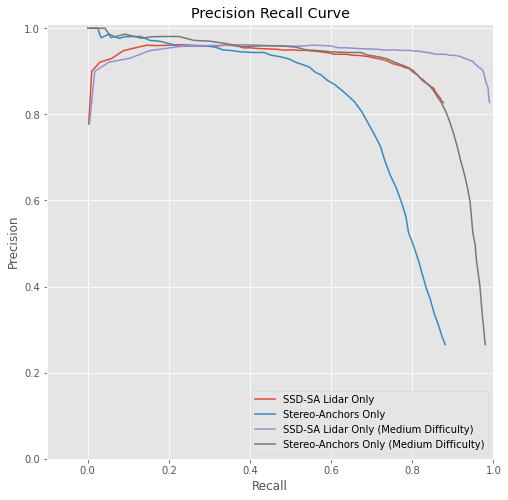}
\caption{Stereo versus LiDAR single-shot ROC curves. The stereo model's degradation on "hard" data indicates relative weakness on heavily occluded, small, or distant objects}
\label{fig:disparitEx}
\end{figure}

\begin{figure}[!htbp]
\centering
\includegraphics[width=1.0\linewidth]{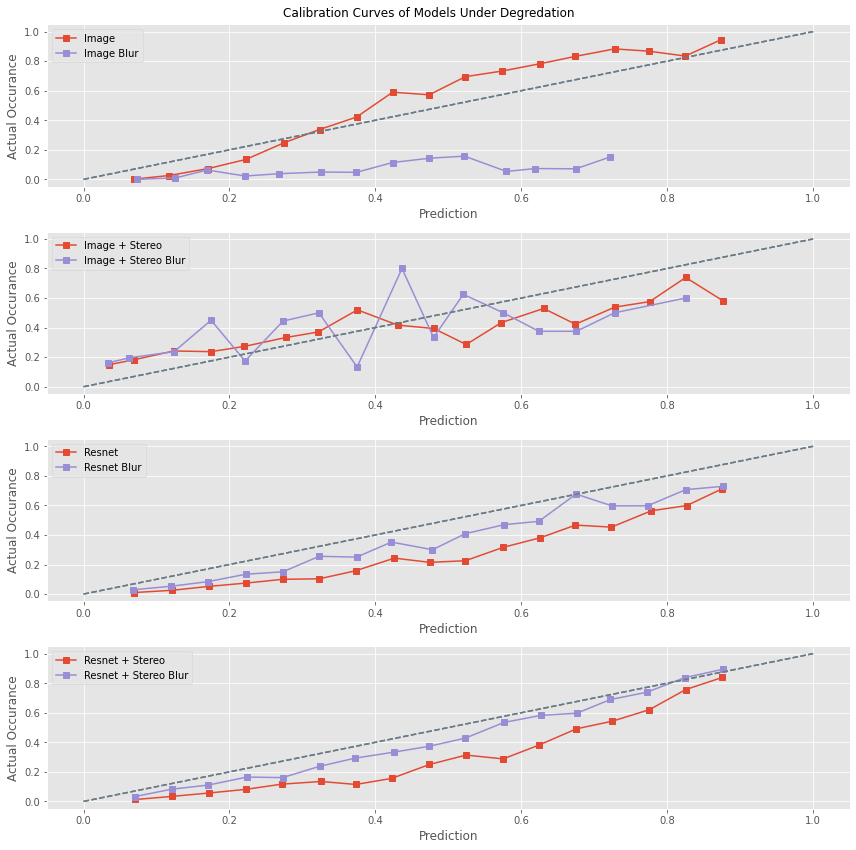}
\caption{Calibration curves for various models before and after image degradation of $\sigma=5$}
\label{fig:calibDe}
\end{figure}

\subsection{Degradation: Performance \& Calibration}

Notably, the stereo-based channel of input allows the model to hold up under certain degradations in the image. In Table 1, we see that the AP for SSD with images drops by $96.9\%$ after a Gaussian pass of $\sigma=5$, but by $39.3\%$ in the model that uses height features. Similarly, though the Brier score of both collapses, the structure of the calibration curve in \ref{fig:calibDe} maintains the qualitative expected shape.

While, typically, image degradation may be expected to result in stereo degradation with traditional block matching, improvements can be made to anticipate this effect \cite{pedone2008blur}, although that is not done here. Even in the case of correlated errors, the calibration derived from imposing additional information in the stereo channel is seen to be more stable and relevant for decision making.

\section{Metric Refinement}

In this section, we make a note of small mistakes used in computing metrics on standard object detection datasets and provide remedy with proof of correctness for relevant components. 

We take the true positive rate (tpr) of an object detector to be the \textit{maximal} true positive classification rate over all possible associations of detections to labels. This definition captures the fundamental tpr goal: measuring how many instances of the primary class the model detects. Existing major object detection metric libraries do not compute tpr along this definition in the cases of overlapping candidates, and do not behave coherently in the presence of filtering criteria. 

\begin{figure}[!htbp]
\centering
\includegraphics[width=0.6\linewidth]{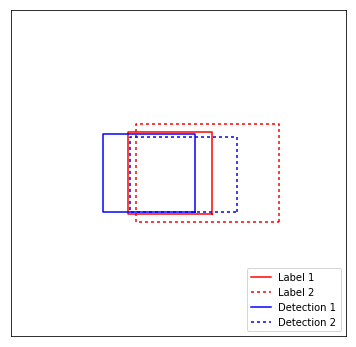}
\caption{Detections on overlapping labels. Despite a possible association that yields two TPs with no FPs, Pascal VOC's "every point interpolation" reports one TP and one FP.}
\label{fig:dets1}
\end{figure}

In Figure \ref{fig:dets1}, the labels (in red) overlap. Some standard object detection libraries fail to accurately associate these detections (in blue) and labels. Such algorithms use a greedy approach to association, iterating sequentially over labels or detections and selecting the candidates of highest overlap. In the case of \ref{fig:dets1}, a greedy algorithm would report one TP, one false negative (FN), and one false positive (FP). While producing a valid classification true-positive rate, this is not accurately capturing the object detection tpr, i.e., there exists an association of labels and detection which, in this case, results in 2 TPs and no FPs or FNs.

\begin{figure}[!htbp]
\centering
\includegraphics[width=0.8\linewidth]{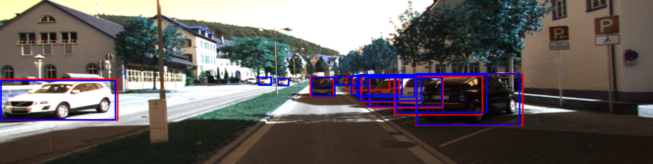}
\caption{KITTI's toolkit mis-associates detections yielding an inaccurate tpr for the SSD-SA model on this frame}
\label{fig:dets2}
\end{figure}

Unfortunately, to exhaustively check all combinations of candidates is intractable, requiring $O(n!)$ complexity in the worst case. The following algorithm provides the mapping of largest tpr while being implemented easily in $O(n^4)$ time.

\subsection{Overview of Algorithm}

\begin{figure}[!htbp]
\centering
\includegraphics[totalheight=3cm]{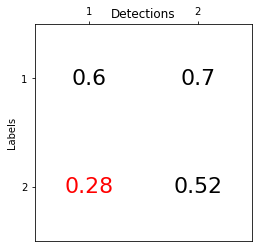}
\includegraphics[totalheight=3cm]{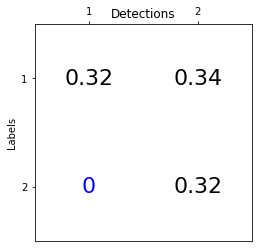}
\caption{(left) Unscaled, unthresholded adjacency matrix for Figure \ref{fig:dets1}, (right) matrix thresholded with $\tau = 0.5$ and scaled}
\label{fig:adj1}
\end{figure}

For any set of detections and labels, take the adjacency matrix $A$ to be constructed such that $A_{i,j}$ is a function of the overlap between detection $i$ and label $j$, thresholded for minimum IoU $\tau$. For Figure \ref{fig:dets1}, the adjacency matrix is shown in Figure \ref{fig:adj1}. In particular,

\[
    A_{i,j} = 
\begin{cases}
    \dfrac{(\text{IoU}_{i,j}+ n)}{2n^2},& \text{if } \text{IoU}_{i,j}\geq \tau\\
    0,              & \text{otherwise}
\end{cases}
\]

An exhaustive search of all candidates would evaluate the permutations of $A$. In the two-by-two case, as in Figure \ref{fig:adj1}, there are only two candidates: $(A_{1,1}, A_{2,2})$, and $(A_{1,2}, A_{2,1})$. This follows directly from the fact that a single label or detection cannot be doubly-associated.

Consider the selector function $f$ for a given adjacency matrix $A$ given as the max of the sum of the permutations of the matrix, i.e.

$$ f(A) = \max_{\sigma\in S_n}{\sum_{i=1}^n{a_{i,\sigma}}} $$

In Figure \ref{fig:adj1}, we evaluate the permutation $\sigma_1$ corresponding to $0.32 + 0.32$ and $\sigma_2$ corresponding to $0.34 + 0$, noting that the first partition has a larger sum.

\begin{lemma}
The maximum sum of permutations in the adjacency matrix $A$ coincides with the maximum true positive (TP) selection
\end{lemma}

First, a note on why, without the scaling factor, $f(A)$ does not coincide in general with the maximum tpr, even though on the example in Figure \ref{fig:adj1} it would yield the correct solution. Indeed, for all $\dim({2})$ matrices and $\tau > 0.5$, $A'$ is sufficient. On the other hand,  $A'$ fails to capture the tpr exactly when any $k$ terms can sum to a lower value than $k-1$ terms, allowing for cases when fewer total terms are selected for a higher sum. Such scenarios are trivial to construct with $\tau = 0.5$ and three detections and labels.

\subsection{Proof of Correctness}

\begin{lemma}$f(A)$ coincides with the maximum tpr
\end{lemma}

For $m$ detections and $n$ labels, we have

$$ A = \begin{bmatrix} a_{1,1} & a_{1,2} & ... & a_{1,n} \\ \vdots & \vdots & \vdots & \vdots \\ a_{m,1} & a_{m,2} & ... & a_{m,n} \end{bmatrix} $$

Assume for simplicity that $m = n$ (if not, take the completion of the matrix with zeros and the proof follows identically. See Figure \ref{fig:badgame} for an example). Assume that $f(A)$ selects too few TPs; in this case, $f(A)$ has returned a permutation $\sigma_i$, when there existed a permutation $\sigma_j$, such that

$$(\sum_{a\in\sigma_i}{a}) > (\sum_{a\in\sigma_j}{a}), \norm{\sigma_j}_0 > \norm{\sigma_i}_0$$

i.e., that the sum of values in one permutation is greater than that in a second, despite the number of elements ($\|\cdot\|_0$) in the second permutation being larger.

Let $k_j = \norm{\sigma_j}_0$ and $k_i = \norm{\sigma_i}_0$ such that $k = k_j - k_i > 0$ by construction.

$$ 
\begin{aligned}
(\sum_{t=0}^{k_j}{\sigma_i[t]}) &> (\sum_{t=0}^{k_j + k}{\sigma_j[t]}) \\
 (k_j)(\dfrac{1}{2n^2} + \dfrac{1}{2n}) &> \frac{(k_j + k)}{2n} \\
 \dfrac{k_j}{2n^2} + \dfrac{k_j}{2n} &> \frac{k_j}{2n} + \dfrac{k}{2n} \\
 \dfrac{k_j}{2n^2} &> \dfrac{k}{2n} \\
 \dfrac{1}{2n} &> \dfrac{1}{2n}
\end{aligned}
$$

Is a contradiction. Not only is $A$ a coinciding solution, it is the minimal coinciding solution. Note that in line 2 we substituted in the bounds for $\sigma$; in line 4, we noted that $k_j <= n$ and $k >= 1$.

$\qed$

We recognize the solution to this problem as the assignment problem, in which we want to find a matching of a particular size in a bipartite weighted graph while maximizing the weights of the edges. We recommend linear programming solutions like the Hungarian algorithm, which runs in $O(n^4)$, because the commonness of its implementation\footnote{In python, this is built-in as \textit{"linear\_sum\_assignment"}}, though $O(n^3)$ solutions exist in the literature. In particular, the adjacency matrix is expected to be largely sparse, and several algorithms opportunize on this structure.

\subsection{Applying Filtering Criteria}

Commonly, a second step in computing metrics is to filter on either the detections or the labels. One may be interested in the AP only on objects greater than some pixel size or closer than some physical distance. Whether performance degrades on objects that are small or far away is especially important to safety cases. KITTI partitions data in to "Easy", "Medium", and "Hard" categories, but only asks that associations outside that group do not penalize the true positive rate; we describe a more robust process.

\begin{figure}[!htbp]
\centering
\includegraphics[width=1.0\linewidth]{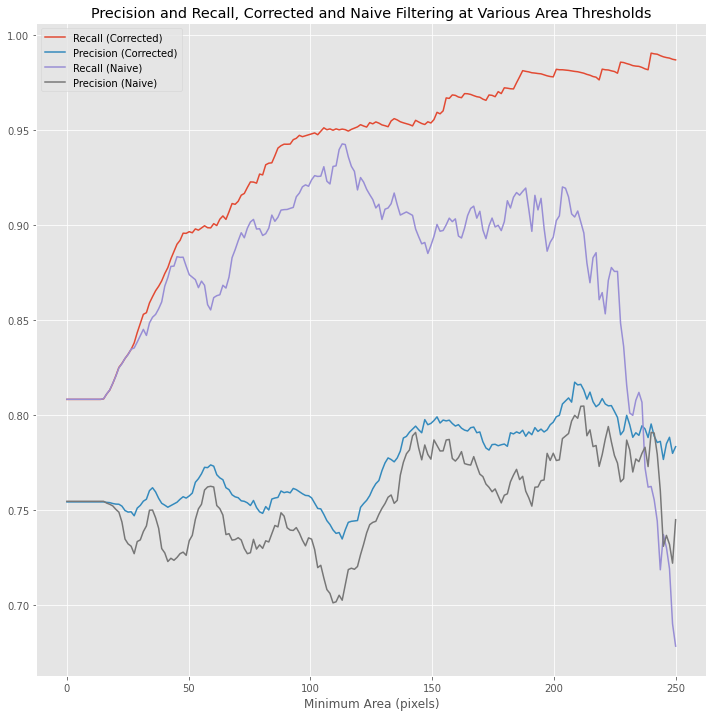}
\caption{Plot of precision and recall for SSD with $\tau=0.25$, minimum confidence $0.25$, when only considering labels and detections of a certain minimum area. Partitioning the data a-priori and then computing a metric on the partition produce very small errors with respect to proposed criteria in this section.}
\label{fig:filterz}
\end{figure}

When evaluating multiple filters, it is helpful to enforce that for a filter to be coherent, it must be "stable", i.e., there should not possibly exist a set of detections and labels such that are more errors on a subset (e.g., the subset with width greater than 40 pixels) than on the total set. Note that aggregate measures or rates - like tpr or recall - can still degrade under stable filters.

One possible approach is to strike from the original set the candidates that do not meet the criteria, and then run the matching algorithm. This results in lack of stability; a simple example occurs when a label is just underneath a threshold but the best matching detection is not. An alternative formulation of the problem would compute tpr conditionally as a constrained optimization. However, the problem is difficult to formulate algorithmically without exhaustively searching candidates, may not be consistent with multiple criteria, and may not guarantee stability.

Instead, a measure on data under a filter can be computed stably as the measure on the subset of the (unfiltered) label-detection pairs (under fixed $\tau$) for which both the detection and label meet the filtering criteria. This formulation has the benefit of being simple and efficient implement. Since only considering the subset of the original associations, one never recomputes the adjacency matrix or the maximal partition, so filtered results can be computed in the minimal possible complexity, $O(n)$. A corollary of the algorithm is that in the case of a multiple solutions (multiple label-detection pairs satisfy the overlap criteria) the algorithm returns the association of highest total overlap. 

Figure \ref{fig:filterz} shows the difference on standard SSD of applying an area filter to labels and detections using the traditional, naive approach and the approach described in this section. Though the difference in recall and precision is small, and only appears in corner cases (these all but disappear, with SSD, if the IoU threshold is the recommended 0.7, which may be why the issue is traditionally ignored), we believe it is important to use accurate definitions of metrics especially in safety-related applications. For this model, the "corrected" metrics tend to be greater than the naive metrics for about $90\%$ of possible area filters up to 200 pixels.

\begin{figure}[!htbp]
\centering
\includegraphics[width=0.6\linewidth]{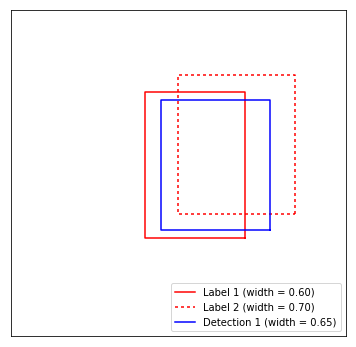}
\caption{Globally sub-optimal metrics from the recommended filtering definition when removing labels with width $< 0.65$}
\label{fig:badgame}
\end{figure}

Although this definition of filters is coherent, it does not guarantee that the tpr is optimal compared to an objective function that included the filtering criteria. Consider the set of detections and labels in \ref{fig:badgame}, assuming we are filtering labels with width $< 0.65$. Note that in this example we take the completion of the original detections by entering null IoU for the missing entries in the $A'$.

In this case, following the proposed algorithm would result in consideration of the subset  {(\textit{Label 2}, $\emptyset$)}, yielding one FN. However, removing \textit{Label 1} under the filter and then computing the mapping would yield a TP. The proposed definition - while stable and coherent - under-represents the number of true positives with respect to that optimality. Accordingly, in \ref{fig:filterz}, we see that no graph moves monotonically with increased filtering, as expected.

Additional investigation may exploit the sparsity implicit in the problem as a further optimization or consider the implicit loss of numerical precision when computing the adjacency matrix as a source of minor numerical instability for high dimensional problems.

\subsection{Computing Brier Score}

\begin{figure}[!htbp]
\centering
\includegraphics[width=1.0\linewidth]{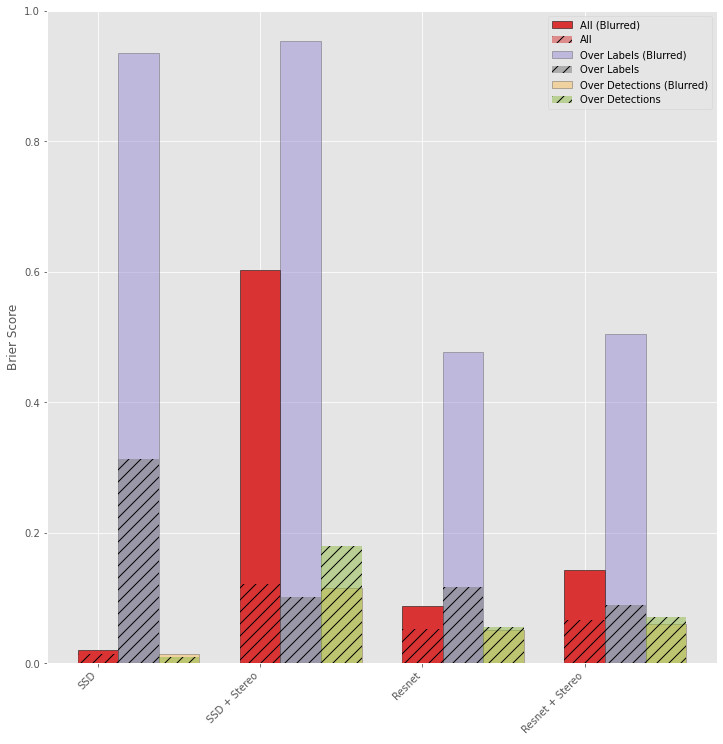}
\caption{Comparison between Brier scores on different subsets of data}
\label{fig:brierly}
\end{figure}

Calibration metrics are less common in the literature, but we offer short consideration of them here. As Figure \ref{fig:brierly} makes clear, the choice of data over which to evaluate Brier score is particularly sensitive. The complexity is introduced because Brier score rewards low-confidence true negatives, and there is a massive ratio of negatives to positives in object detection. Although the default method of computing Brier score is to consider the set of all labels and detections for a frame, it  produces highly counterintuitive results that we do not recommend for use. For a particular image, a model that produces no detections on a negative region makes no contribution to the cumulative Brier score, but a different model that produces a very low confidence positive object detection in a negative region will be rewarded in terms of its contribution to the Brier score. This result should be avoided, since the first model is clearly better calibrated. On the other hand, only examining over the set of \textit{detections} leaves out FNs, one of the core measures of object detectors. Thus, when reporting Brier score, we report over the set of labels, precluding miscalibration on FPs but providing the most honest assessment of performance. Note that this decision has a much greater effect on results than the decision in the rest of Section 4, in that it actually flips the conclusion one might derive when evaluating in the usual way. Thus, we present all of the possible scores in this section, but advocate for the score that matches metric intuition in unbalanced cases.

{\tiny
\bibliographystyle{IEEEtran}
\bibliography{egbib}
}

\end{document}